\documentclass[PHM, 2021]{PHMSociety}

\usepackage{graphicx}
\usepackage{amsmath}
\usepackage{multirow}
\usepackage{tikz}
\usepackage{booktabs}

\long\def\/*#1*/{}

\begin{document}

\title{A stacked deep convolutional neural network to predict the remaining useful life of a turbofan engine}


\author{%
	David Solís-Martín \authorNumber{1,2} , Juan Galán-Páez\authorNumber{1,2} , and Joaquín Borrego-Díaz\authorNumber{2} 
}

\address{
	\affiliation{{1}}{Datrik Ingellicence S.A., Seville, Spain}{ 
		{\email{david.solis@datrik.com}},
		{\email{juan.galan@datrik.com}}
		} 
	\tabularnewline 
	\affiliation{2}{Department of Computer Science and Artificial Intelligence, Seville University, Spain}{ 
		{\email{dsolis@us.es}},	{\email{juangalan@us.es}},
		{\email{jborrego@us.es}}
		} 
}

\maketitle

\phmLicenseFootnote{David Solís-Martín}

\begin{abstract}

This paper presents the data-driven techniques and methodologies used to predict the remaining useful life (RUL) of a fleet of aircraft engines that can suffer failures of diverse nature. The solution presented is based on two Deep Convolutional Neural Networks (DCNN) stacked in two levels. The first DCNN is used to extract a low-dimensional feature vector using the normalized raw data as input. The second DCNN ingests a list of vectors taken from the former DCNN and estimates the RUL. Model selection was carried out by means of Bayesian optimization using a repeated random subsampling validation approach. The proposed methodology was ranked in the third place of the 2021 PHM Conference Data Challenge.
\end{abstract}

\section{Introduction}

Remaining useful life (RUL) is a cross-disciplinary field that belongs to the wider field of Prognostics and Health Management (PHM). This discipline studies the system behavior during its lifetime, that is, from the last check or maintenance performed on it until the system fails or the degradation of the system performance exceeds a certain threshold. Therefore, RUL enables to estimate the future reliability and scans the degradation of the system along the time \cite{peng2018switching}. The final goal is to estimate the remaining time until such failure occurs or a degradation level is reached, given the system working condition at any point of the system's lifetime.

Two main methodologies for RUL estimation can be found in the literature: model-based methods and data-driven methods. Model-based methods establish a degradation model based on physics and statistics to predict the degradation trend of the system. Model-based methods require a strong knowledge on the physical behavior of each specific component and failure type, thus it is difficult to develop models for the current increasingly complex systems \cite{zhao2017remaining}. Moreover, some assumptions need to be taken on the models, thus those can be biased. For all these reasons, model-based methods have a limited prediction performance and have to be designed ad hoc for each different type of machinery.

On the other hand, data-driven methods use machine learning algorithms to learn the degradation trend of the system using historical condition monitoring data. In contrast to model-based methods, prior expertise is not required. Their application can be straightforward when enough data of quality are available. 


Among the existing data-driven methods, Deep Learning (DL) is one of the most popular and promising fields of study. During the last decade, the use of DL techniques has surged significantly. Especially in complex tasks with high-dimensional nonlinear data. DL has had an enormous success in image processing, natural language processing, and signal processing. For this reason, it is not unexpected that DL based approaches are also widespread within the context of Prognostics and Health Management (PHM) research. One type of network that has been used to deal with sequence data is the deep convolutional neural network (DCNN) \cite{lecun1999object}. For example, in \cite{zhao2017convolutional} the authors applied DCNN to time series classification over a variety of datasets. Within of the context of PHM, \cite{babu2016deep} and \cite{li2018remaining} authors have applied DCNN to RUL prediction of aircraft turbofan engines.

\section{Data and problem description}

The Commercial Modular Aero-Propulsion System Simulation (CMAPSS) is a modeling software developed at NASA. It was used to create the previous CMAPSS dataset \cite{saxena2008damage}. That first dataset was built by only taking samples from a system that is already degraded. "\emph{Therefore, the onset of the fault cannot be predicted; only the evolution of the fault can}" \cite{arias2021aircraft}. 

\begin{figure}[t]
\centering
\includegraphics[scale=.3]{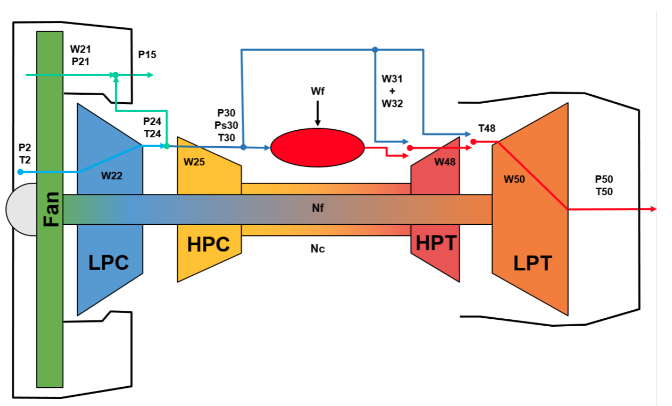}
\caption[Schematic of the model used in CMAPSS.]{Schematic of the model used in CMAPSS \cite{arias2021aircraft}.}
\label{fig:cmapss_model}
\end{figure}

A new dataset, named N-CMAPSS, has been created providing the full history of the trajectories starting with a healthy condition until the failure occurs. A schematic of the turbofan model used in the simulations is shown in Figure \ref{fig:cmapss_model}. All rotation components of the engine (fan, LPC, HPC, LPT, and HPT) can be affected by the degradation process. This model is defined as a nonlinear system denoted by:

\begin{equation}
\left [x_s^{(t)}, x_v^{(t)} \right ] = F \left ( w^{(t)}, \theta^{(t)} \right )
\label{eq:cmapss_system}
\end{equation}

where $x_s$ are the physical properties, $x_v$ are the unobserved properties, $w$ is the scenario-descriptor operating conditions and $\theta$ are the health model parameters.

\subsection{Data}

The dataset provided for the challenge consists of data from 90 simulated flights (extracted from the N-CMAPSS\footnote{The N-CMAPSS dataset can be downloaded from the \href{https://ti.arc.nasa.gov/tech/dash/groups/pcoe/prognostic-data-repository/\#turbofan-2}{\underline{NASA data repository}}} \cite{arias2021aircraft}). Seven different failure modes, related to flow degradation or subcomponent efficiency that can be present in each flight have been defined. The flights are divided into three classes depending on the length of the flight. In class 1, fall flights with a duration from 1 to 3 hours, class 2 is composed of flights between 3 and 5 hours, and class 3 contains flights that take more than 5 hours. Each flight is divided into cycles, covering climb, cruise, and descend operations.

\subsection{Problem definition}

The problem revolves around the development of a model $\mathcal{G}$ capable of predicting the remaining useful life $Y$ of the system, using the sensor outputs $X_s$, the scenario descriptors $W$ and auxiliary data $A$. The different variables available to estimate the RUL of the system are described in table \ref{table:variables}. The former is an optimization problem that can be denoted as:

\begin{equation}
argmin \: \sum \mathcal{S} \left ( y - \hat{y} \right )
\end{equation}

where $y$ and $\hat{y} = \mathcal{G}(x_s, w, a)$ are, respectively, the expected and estimated RUL. $\mathcal{S}$ is a scoring function defined as the average of the root-mean-square error (RMSE) and the NASA's scoring function ($N_s$) \cite{saxena2008damage}:

\begin{equation}
\mathcal{S} = 0.5 \cdot RMSE + O.5 \cdot N_s
\label{eq:score}
\end{equation}

\begin{equation}
N_s = \frac{1}{M} \sum exp(\alpha | y - \hat{y} |) - 1
\end{equation}

being $M$ the number of samples and $\alpha$ equal to $\frac{1}{13}$ in case $\hat{Y}  < Y$ and $\frac{1}{10}$ otherwise.

\begin{table}[]
\caption{Variable description.}
\begin{tabular}{llll}
\hline
Symbol & Set & Description                     & Units \\ \hline
alt    & $W$   & Altitude                        & ft    \\
Mach   & $W$   & Flight Mach number              & -     \\
TRA    & $W$   & Throttle-resolver angle         & \%     \\
T2     & $W$   & Total temperature at fan inlet  & ºR    \\
Wf     & $X_s$ & Fuel flow                       & pps   \\
Nf     & $X_s$ & Physical fan speed              & rpm   \\
Nc     & $X_s$ & Physical core speed             & rpm   \\
T24    & $X_s$ & Total temperature at LPC outlet & ºR    \\
T30    & $X_s$ & Total temperature at HPC outlet & ºR    \\
T48    & $X_s$ & Total temperature at HPT outlet & ºR    \\
T50    & $X_s$ & Total temperature at LPT outlet & ºR    \\
P15    & $X_s$ & Total pressure in bypass-duct   & psia  \\
P2     & $X_s$ & Total pressure at fan inlet     & psia  \\
P21    & $X_s$ & Total pressure at fan outlet    & psia  \\
P24    & $X_s$ & Total pressure at LPC outlet    & psia  \\
Ps30   & $X_s$ & Static pressure at HPC outlet   & psia  \\
P40    & $X_s$ & Total pressure at burner outlet & psia  \\
P50    & $X_s$ & Total pressure at LPT outlet    & psia  \\
Fc     & A   & Flight class                    & -     \\
$h_s$    & A   & Health state                    & -    
\end{tabular}
\label{table:variables}
\end{table}

\/*

*/

\section{Proposed methodology}

This section introduces the different phases or processes in which the proposed methodology can be decomposed.

\subsection{Data normalization}

The 20 variables used to develop the different models (see table \ref{table:variables}) have different scales. For that reason, directly feeding the proposed networks with such data will slow down the learning and convergence of the models. Hence, a data normalization step, before the training stage, is required to homogenize the variables into a common scale. More precisely, the standard normalization is used in this paper:

\begin{equation}
    x'_f = \frac{x_f - \mu_f}{\sigma_f}
\end{equation}

where $x_f$ is the data of a feature $f$, and $\mu_f$ and $\sigma_f$ are its mean and standard deviation, respectively. Note that the mean and variance are computed for each training cross-validation set and are also used to normalize the validation sets.

\subsection{Time window processing}

\begin{figure}[t]
\centering
\includegraphics[scale=.2]{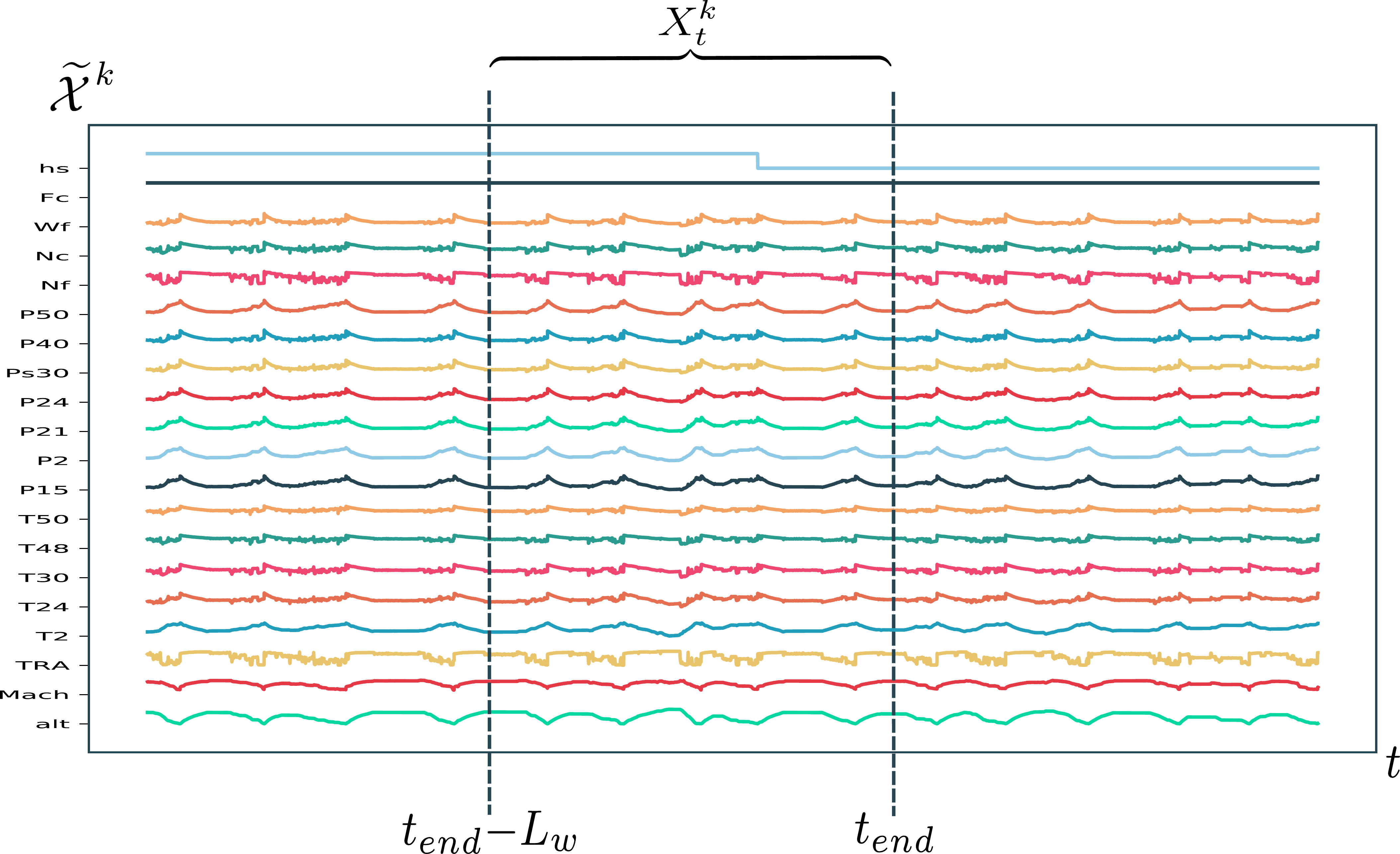}
\caption{Sliding window.}
\label{fig:window}
\end{figure}

After the data normalization, the inputs of the network are generated using a sliding time window through the normalized data. The size of the window, denoted as $L_w$, is a parameter to be selected during the model selection stage. The inputs generated can be expressed as $X^k_t = [\widetilde{\mathcal{X}}^k_{t_{end} - L_w}, ..., \widetilde{\mathcal{X}}^k_{t_{end}}]$ (see Figure \ref{fig:window}) and the paired ground-RUL label is $Y_t$. With this approach, for each unit, $T^k - L_w$ samples will be considered. Where $T^k$ is the total run time in seconds of each unit.

It is common to set a constant RUL at the beginning of the experiment \cite{li2020remaining}, considering a healthy condition of the system during that early stage. In this work, this approach has not been followed since this kind of assumptions can introduce a bias in the model. Instead, the ground-RUL label has been set as a linear function of cycles from the RUL of each unit $Y^k_t = TUL^k -  C^k_t$, where $TUL^k$ is the total useful life of the unit $k$ in cycles and $C^k_t$ is the number of past cycles from the beginning of the experiment at time $t$.

\begin{figure}[t]
\centering
\includegraphics[scale=.12]{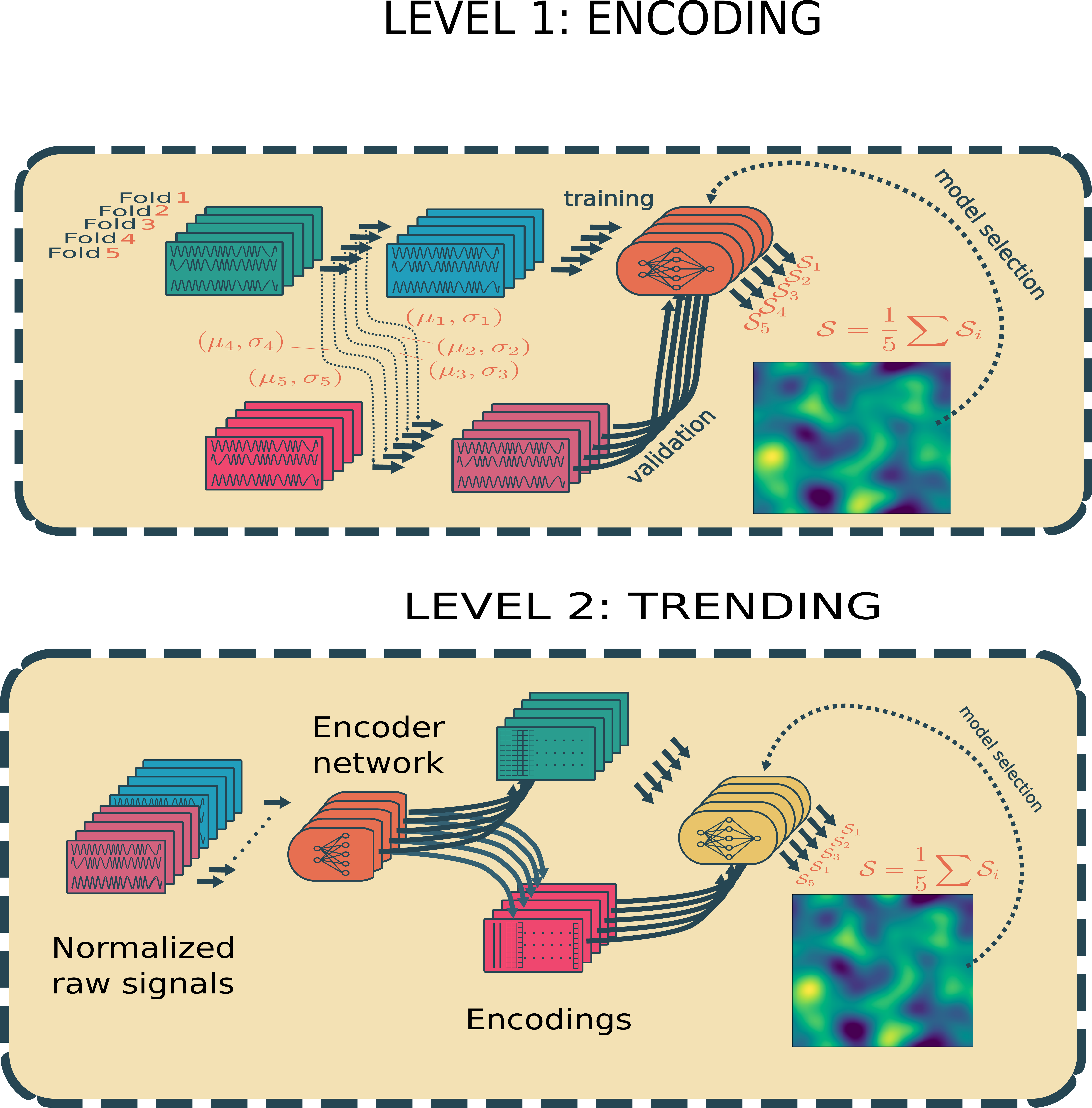}
\caption{Methodology}
\label{fig:method}
\end{figure}

\subsection{Level 1 and level 2 model: Convolutional Neural Networks}

The modeling phase consists of two levels (see Figure \ref{fig:method}). In the first level, the goal is to find a good model to produce a time-windowed encoding of the raw input signals. This encoding is needed due to the high dimension of the input data. Another goal of this encoding step, besides the dimension reduction of the input, is to remove as much noise as possible. Such an encoding is used as the level 2 model input. The goal of this second level model is to provide an estimation of the RUL. 


For both level models, Deep Convolutional Neural Networks (DCNN) have been selected. DCNN is a specialization of fully connected networks (FCN) that have been widely applied in image processing, natural language processing, and speech recognition with great success. The CNN uses parameter sharing and subsampling to extract feature maps with the most significant local features. The main operations in a DCNN are convolution and polling 
. The convolution operation implements parameter sharing and local receptive fields. The equation of the convolution is as follows:

\begin{equation}
S\left [ i,j \right ] = \sum_m \sum_n I \left [ i + d*n,j + d*m \right ] K \left [ n, m \right ]
\end{equation}

where I is an input matrix, K is the kernel matrix (or convolution's parameters) of size $n$ x $m$, $d$ is the dilation rate, and $S$ is the result of the convolution, called feature map. The kernel matrix is slipped across the input matrix looking for a pattern present in any place of it. Thereby, comparing with a FCN, the number of weights is reduced and, additionally, the overfitting chance. The pooling operation performs a downscaling by applying a statistic operation to each region of the input, which was previously divided into rectangular pooling regions. The pooling operation serves a few porpouses: reduces the computational requirements for the upper layer since the feature maps are downscaled, reduces the number of connections (parameters) for the upper fully connected layers, provides a form of spatial transformation invariance and helps mitigating the overfitting risk.

\subsection{Cross validation}

\begin{figure}[t]
\centering
\includegraphics[scale=.25]{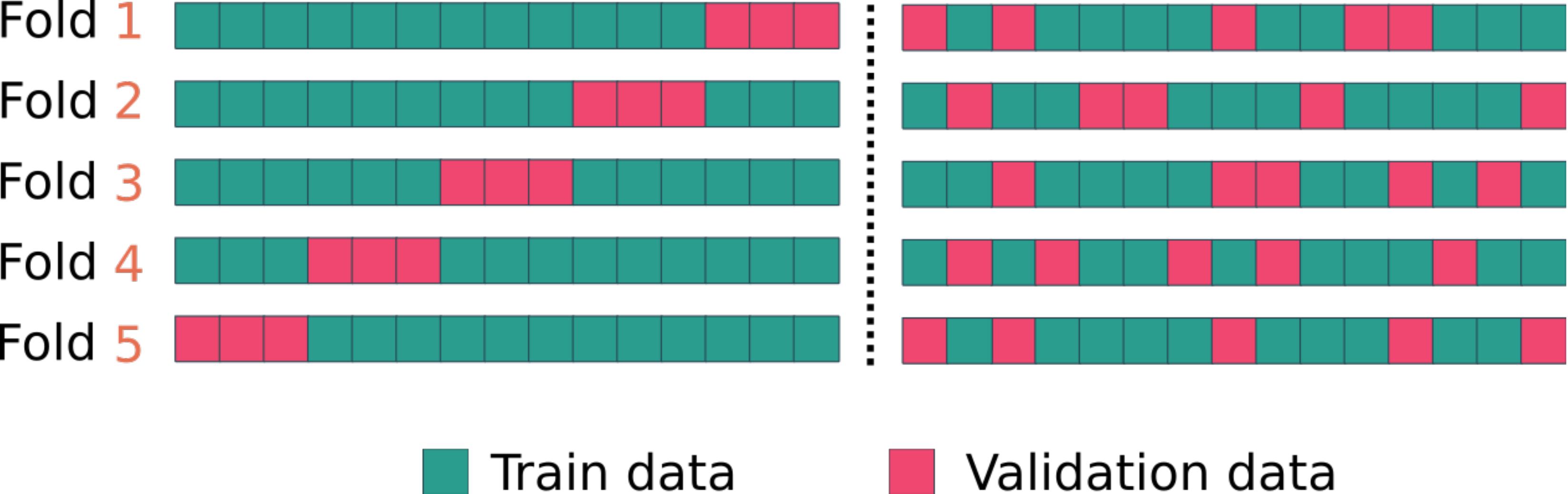}
\caption{$k$-fold cross validation strategy (left), and $k$-fold random sampling strategy (right).}
\label{fig:cv}
\end{figure}

Neural networks have a high number of parameters to be adjusted. Additionally, the size of the sliding window has to be optimized too. To deal with the overfitting problem during the model selection phase, a bunch of validation techniques can be found in the literature. The classical single hold-out strategy is discarded since it can easily overfit the hold-out set and lead to poor generalization results. The $k$-fold cross-validation has the disadvantage that the size of the validation set decreases as $k$ (the number of folds) is increased. There exist alternatives as the $k$-fold repeated random subsampling validation that overcome this issue (see Figure \ref{fig:cv}). The later strategy has been selected in this work, with $k=5$ and a validation fold size of 30\% of the training set, that is, 27 random units from a total of 90.

\subsection{Model Selection}


The goal of the model selection phase is to obtain an optimal parameter configuration for each model. To this aim, Bayesian optimization has been selected as the optimization strategy. The models were trained using the RMSE as the loss function, since the NASA score is not differentiable. However, the loss function used in the Bayesian optimization to decide the next set of model parameters to be tested, is the score $\mathcal{S}$, defined in equation (\ref{eq:score}). 


The well-known \emph{early stopping} method, with a patience factor of 8 epochs, is used as a signal to finish the model training process. Hence, the training of the model will be stopped if the training loss on the validation fold has not been improved during the last 8 training epochs. Additionally, the learning rate is decreased by a factor of 0.1 when no improvement is seen on the validation loss in the last 3 training epochs.

\section{Experiments and results}

This section describes the settings and parameters used to find the best models. Then the results of each level are compared and the final solution is described. Finally, this approach is applied to the hidden test data.

\subsection{Level 1: encoding}

\begin{figure}[t]
\centering
\includegraphics[scale=.09]{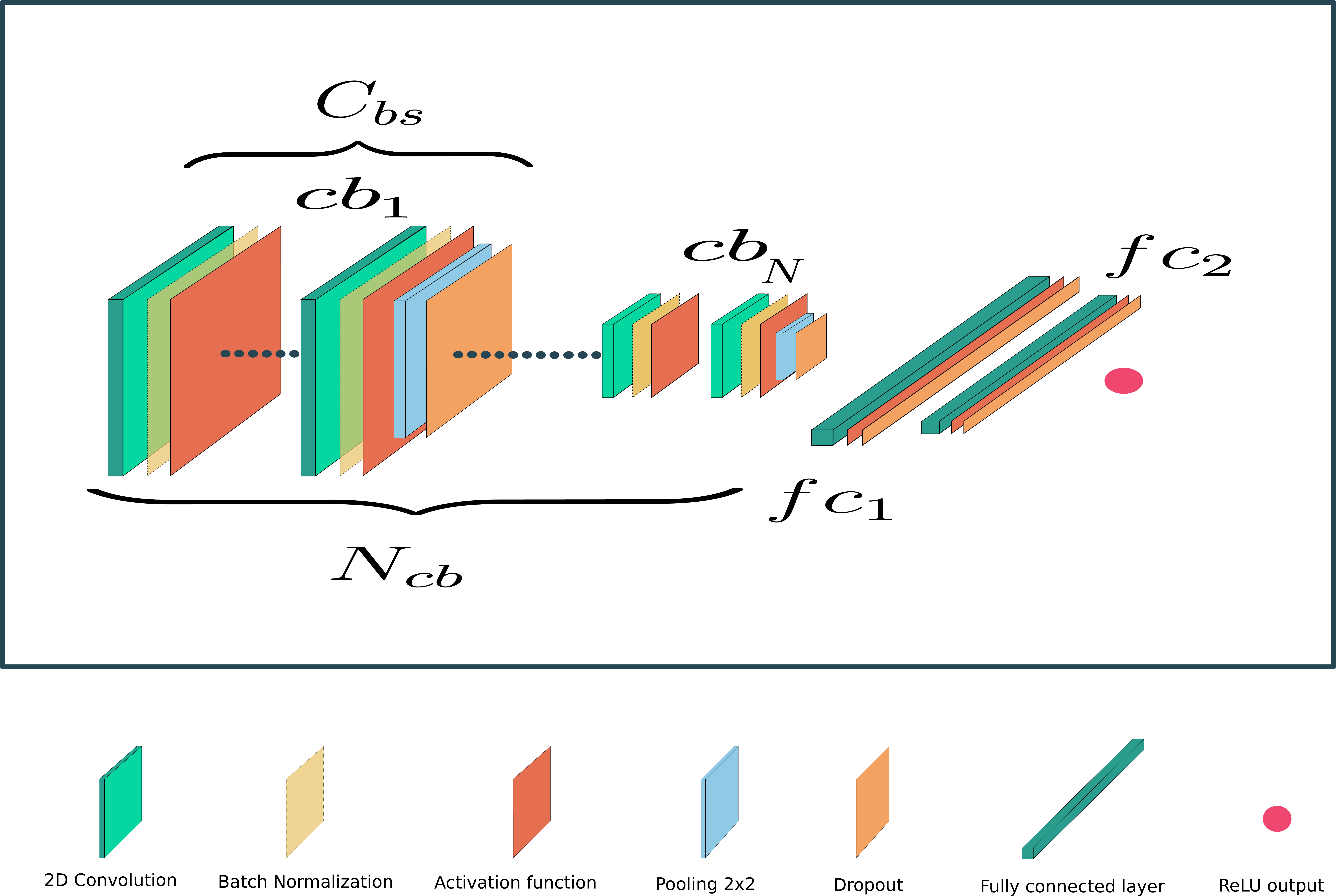}
\caption{Level 1 networks architectures tested.}
\label{fig:architectures}
\end{figure}

The encoding model at level 1 of the stacking aims to carry out a dimension and noise reduction on the raw signals. To achieve this task, DCNN architecture was used. The classical DCNN architecture, which is shown in Figure \ref{fig:architectures}, can be divided into two parts. The first one is a stacking of $N_b$ blocks composed of convolution and pooling layers. The goal of this first part is to extract characteristics potentially useful for the task. The second part consists of fully connected layers and in this case will perform the regression of the RUL.

To obtain an optimal hyperparameter configuration, Bayesian optimization was executed during 100 iterations. The first 10 iterations, consist in models with randomly selected hyperparameters. These first 10 random points of the model hyperparameter search space are used by the Bayesian process to create the initial estimation of the error surface. After these 10 iterations, the optimization process applies Bayesian rules to select the most promising point (model hyperparameter) to be tested. Table \ref{table:param_ranges} summarizes the input parameter ranges and the best parameter configuration found for the  L1 model.

Figure \ref{fig:predl2} (top) shows the predictions for 4 units made by the DCNN  model. The confidence intervals have been computed as $[\mu_p - 3\sigma_p, \mu_p + 3\sigma_p]$ where $\mu_p$ and $\sigma_p$ are the mean and variance, respectively, of the cross-validation predictions. As it can be observed in the charts, the models are more reliable when the failure occurs earlier. In the first cycles, the prediction seems to be almost constant, until the model is able to catch the descending trend. 

\subsection{Level 2: RUL estimation}




The aim of the level 2 model of the stacking is to perform the final RUL predictions. The input of this model is the encoding of the raw signal generated by some of the level 1 models. Therefore, a first step is needed to generate the encoding for the train and test sets per fold. The encodings are extracted from the second full connected layer of 100 neurons.

\begin{table}[]
\centering
\caption{Parameter ranges used in the Bayesian optimization and values of the best models for level 1 (L1) and level (L2)}
\begin{tabular}{lcrr}
\specialrule{2pt}{1pt}{1pt}
\textbf{Patameter}             & \textbf{Range or value}  & \textbf{L1}    & \textbf{L2}    \\ 
\specialrule{2pt}{1pt}{1pt}
$L_w$                       & [100, 500] & 161   & -       \\      
\specialrule{1.2pt}{1pt}{4pt}
$B_s$                       & {32,64}    & 116   & 31           \\   \hline
$C_{bs}$                    & [2,4]      & 4     & 4                    \\ \hline
$N_{cb}$                    & [2,4]      & 4     & 4           \\ \hline
$l_1$                       & [0, 1e-3]  & 7.23e-4  & 6.96e-4          \\   \hline
$l_2$                       & [0, 1e-3]  & 0        & 1.73e-5  \\  \hline
$l_r$                       & [1e-5, 1e-3] & 0.001    & 5.53e-4    \\   \hline
dropout                     & [0, 0.9]  & 0.13  & 0.21     \\   \hline
$fc_1$                      & [64, 256] & 100 & 247                 \\ \hline
$\sigma_{conv}$             & \{tanh, ReLU,  &  &    \\ 
                            &  leakyReLU\} & tanh & tanh   \\ \hline
$d_{rate}$                  & [1, 10]                     & 2  & 2     \\ \hline
$K_s$                       & \{(3,3), (10,1), &     &    \\ 
                            & (10,3)\} & (10,1)     & (10,1)   \\ \hline
$\sigma_{fc}$               & \{tanh, ReLU, &  &         \\
                            &  leakyReLU\} & leakyReLU & leakyReLU        \\ \hline
$\sigma_{output}$           & ReLU                      &  ReLU &  ReLU    \\ 
\specialrule{1.2pt}{1pt}{4pt}
$fc_2$                      & [100, 1000] &  - &     105              \\ \hline
$channels$                      & [1, 3] &  - &     3              \\ \hline
$step$                      & [64, 256] &  - &     989              \\ 
\specialrule{2pt}{1pt}{4pt}
\#Net params    & &  1,514,016 & 2,575,449         \\ \hline
RMSE            &   &    10.46  & 6.24    \\ \hline
MAE             &  &    7.689   & 4.27             \\ \hline
NASA score      &    &  2.13 & 0.64                \\ \hline
CV \: $\mathcal{S}$ score &  & 6.30  & 3.44 \\ \hline
std($\mathcal{S}$)        &  &   0.37 & 0.63  \\ \hline
Ensemble  &  &   &  \\
$\mathcal{S}$ score &  & -  & 2.95 \\
\end{tabular}
\label{table:param_ranges}
\end{table}


Each sample input for each candidate level 2 model will be generated using a sliding window over the encodings. Since the encodings are a dimension reduced version of the raw signal, the level 2 model can expand the receptive field of the model and learn from the trend along the time. The parameter \emph{step} defines the gap in seconds between encodings. 

\begin{figure}[t]
\centering
\includegraphics[scale=.13]{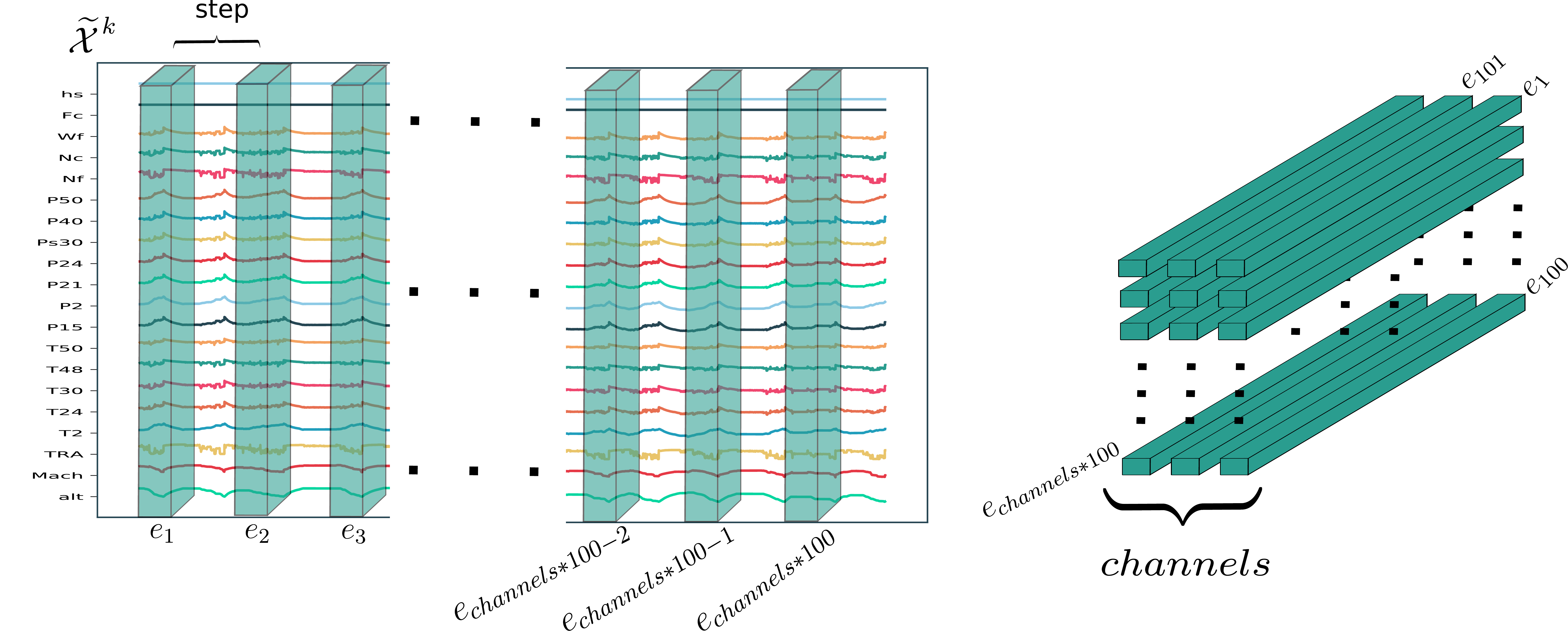}
\caption{Sliding window for l2 models. Left: window over the raw signals to generate the encodings $e_i$. Right: encoding input shape for the DCNN.}
\label{fig:window2}
\end{figure}

Figure \ref{fig:window2} shows how the inputs for the models are obtained. The concept of image channel is used to compose the input. Each channel has 100 vector encodings. Therefore, the total number of encodings in the input is $100 \cdot chanells$. 


The level 2 model is also a DCNN. The cross-validation schema and parameter ranges to be optimized are the same considered in the level 1 model. Addicionaly, three more parameters need to be considered in this L2 model optimization, namely $fc_2$, $step$, and $channels$ (see Table \ref{table:param_ranges}).

Table \ref{table:param_ranges} summarizes the best parameter configuration found for the level 2 model, which is quite similar to that of level 1. This could be due to that the level 1 hyperparameters are provided to the optimization process as seed. There exist differences only in the training and regularization parameters. The rest of the architecture is identical. Regarding the additional parameters, it is interesting to note that the value selected for the parameter \textit{step} is 989. Since the level 1 model provides a fairly good estimation of the RUL, the level 2 model only needs a set of sparse encodings to capture the trend and improve the RUL. The cross-validation score of this model is 3.44, while the ensemble score (that is, taking the average of predictions before computing the score), is 2.95. This gap between the cross-validation scores and ensemble predictions means that the predictions of the models have a good level of uncorrelation.

\begin{figure}[t]
\centering
\includegraphics[scale=.21]{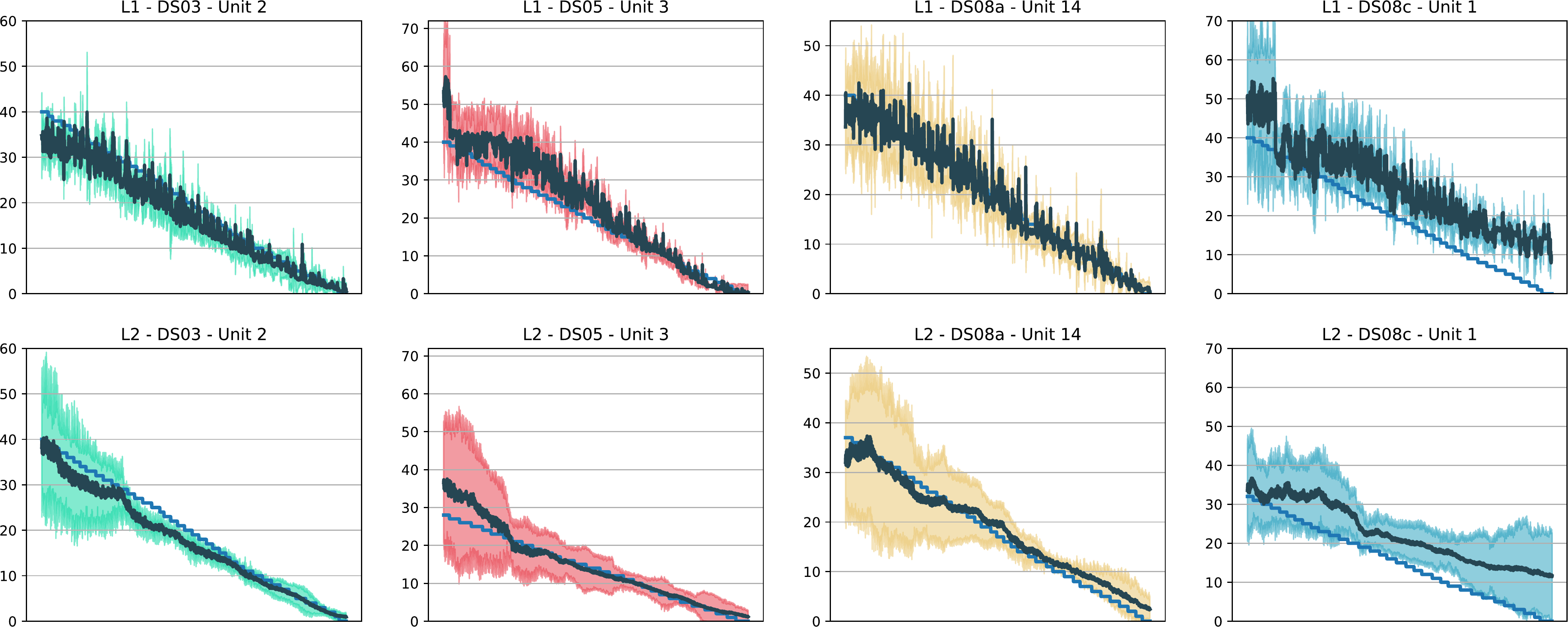}
\caption{Prediction and confidence interval for 4 units the level 1 model (top) and the level 2 model (bottom).}
\label{fig:predl2}
\end{figure}

\begin{figure}[t]
\centering
\includegraphics[scale=.4]{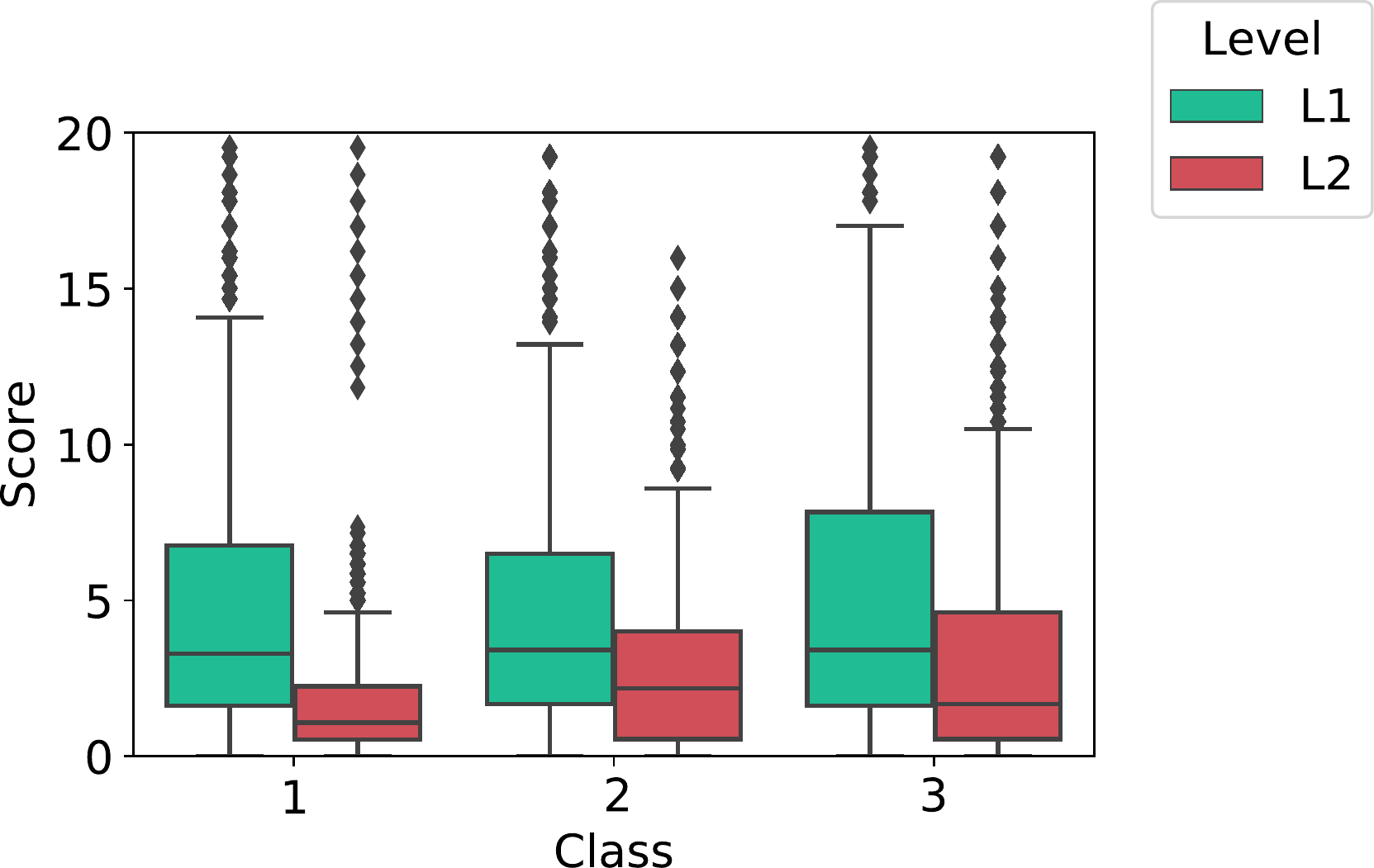}
\caption{Performance of the models per class.}
\label{fig:score_klass}
\end{figure}

Figure \ref{fig:predl2} (bottom) shows the predictions and confidence intervals of the level 2 models. Comparing these with the level 1 models (Figure \ref{fig:predl2} top), one can see that the predictions of the stacked models have been smoothed and improved. Figure \ref{fig:pred_score} presents the relation between the ground truth RUL and the score obtained by each stacked model. 
Finally, figure \ref{fig:score_klass} shows the performance of each  model in each class. Note that the shorter are the flights, the better the model performs. In the same way, the improvement in the level 2 model performance is higher for shorter flights.

\subsection{RUL prediction on test data}

\begin{figure}[t]
\centering
\includegraphics[scale=.18]{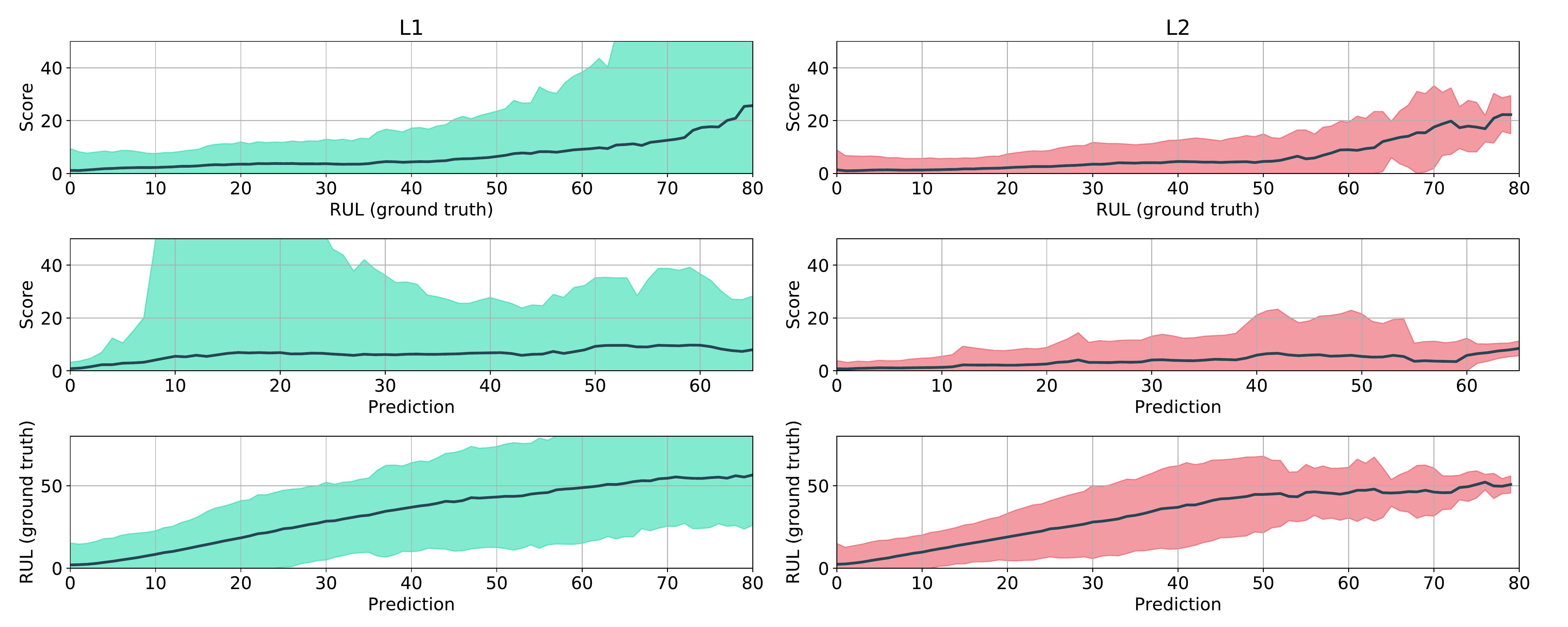}
\caption{The chart shows the relation beetween  RUL and the score (top), the prediction and the score (middle) and the prediction and RUL (bottom).}
\label{fig:pred_score}
\end{figure}

A common approach to obtain the final model consists in training it using all available data. In this approach, a number of training epochs must be selected, thus the validation data used during the model selection would also be part of the final training set. To select the number of epochs to train, the mean value among the best epoch obtained for each fold is considered. Usually, it would be necessary to save part of the training set as the final test set to validate the performance of the model. Hence, the number of samples in the training/validation set is reduced. 
A different approach has been used in the proposed methodology, which consists in using all models trained per fold as an ensemble. This approach has two main advantages. The first one is that it is possible to obtain a confidence metric of our final ensemble model performance by means of cross validation. The second one, related to the first, is that the predictions per sample could be used to create a confidence interval of the final mean prediction as it is shown in the Figure \ref{fig:predl2} and \ref{fig:pred_score}.

The hidden validation set to be scored by the competition committed was composed of 38 units. The predictions generated by the final solution are: 22, 21, 19, 24, 13, 18, 12, 22, 11, 6, 22, 24, 19, 10, 19, 21, 20, 17, 20, 26, 19, 12, 11, 13, 9, 37, 25, 4, 18, 25, 18, 14, 11, 12, 22, 10, 19, 19, which scored 3.651. This score is very close to the cross-validation score. An estimation of the score can be obtained by using the prediction/score distribution (Figure \ref{fig:pred_score} bottom). However, the ensemble score is 2.95, which is far from the real score obtained. This means that there is room for improvement by reducing this overfit.



\section{Conclusions}

This work has exposed a robust methodology to estimate RUL without the need of expert knowledge on the system studied. The success of this methodology has been demonstrated with the results obtained in the 2021 PHM Conference Data Challenge (ranked in the third place). The goal of the challenge was to predict RUL in turbogan aircraft engines, using data generated with a CMAPSS simulator.

In this work, the process of obtaining the final solution is divided into two learning stages. In the first one, an encoding of the raw data is learned and used as input of the second learning stage to obtain the final model capable of estimating RUL. The finals results show that using one of the most basic architectures found in the literature is enough to achieve an excellent outcome. 

A number of possible improvements of the current solution will be the target of future research: 1) Smoothing the RUL among cycles could smooth the error space, thus helping in the learning process. 2) In the proposed methodology, the inputs having a number of cycles lower than that required by the network, have been excluded from the training process. In the case of the validation set, these inputs were filled with the earliest encoding. It is expected that training the networks applying the same filling will help reducing the overfitting. 
3) Another possible improvement could be to balance the training folds so each failure type is properly represented in each validation set. 4) Train the level 2 model with a random gap between encodings. This allow to expand the training set, have an additional mechanism to compute the confidence intervals, and increase the number of predictions to calculate the final averaged RUL. 5) Finally, it would be interesting to study how predictive are the class and the health state features.

This work has been developed with a focus on reproducible research. To this aim, the parameters obtained and the parameter ranges used for model selection have been well described. In addition, the source code to train the models and validate the results can be found in the source code repository \href{https://github.com/DatrikIntelligence/Stacked-DCNN-RUL-PHM21}{https://github.com/DatrikIntelligence/Stacked-DCNN-RUL-PHM21}.

\section*{Acknowledgment}

This work has been supported by Grant PID2019-109152GB-I00/AEI/10.13039/501100011033 (Agencia Estatal de Investigación), Spain and by the Ministry of Science and Education of Spain through the national program “Ayudas para contratos para la formación de investigadores en empresas (DIN2019)”, of State Programme of Science Research and Innovations 2017-2020.

\section*{Nomenclature}

\begin{tabular}{ l  l }
	$\mathcal{X}$			& Raw signals dataset \\
	$\widetilde{\mathcal{X}}$	& Normalized signals dataset\\
	$\mathcal{Y}^k_t$	    & RUL of a unit $k$ at time $t$ \\
	$\widetilde{\mathcal{X}}^k_t$			& Normalized signals at time $t$ of a unit $k$ \\
	$L_w$                   & Length of the sliding window \\
	$\widetilde{X}^k_t$     & Normalized signals of a unit $k$ \\
	\:                      & between $t - L_w$ and $t$ \\
    $TUL^k$                 & Total useful life of the unit $k$ \\
	$C^k_t$                 & Cycle number of the unit $k$ at time $t$     \\ 
	$C_bs$                  & Convolution block size \\
	$N_cb$                  & Number of convolution blocks \\
	$d_{rate}$              & Dilation rate of a convolution \\
	$fc_x$                  & Neurons in the $x$-th fully connected layer\\
	$\sigma_{conv}, \sigma_{fc}, \sigma_{out}$ & Activation function of a convolution layer, \\  
	\:                                            & fully connected layer and the output layer \\ 
	\:                                            & of a network. \\
	$K_s$                   & Kernel size of a convolution \\
	$l_1, l_2$                   & Weights for $L_1$ and $L_2$ regularization  \\
	$l_r$                   & Learning rate  \\
	$B_s$                   & Batch size \\
 \end{tabular}

\bibliographystyle{apacite}
\bibliography{ijphm}

\begin{thebibliography}{}

\bibitem [\protect \citeauthoryear {%
Abadi%
\ \protect \BOthers {.}}{%
Abadi%
\ \protect \BOthers {.}}{%
{\protect \APACyear {2016}}%
}]{%
abadi2016tensorflow}
\APACinsertmetastar {%
abadi2016tensorflow}%
\begin{APACrefauthors}%
Abadi, M.%
, Barham, P.%
, Chen, J.%
, Chen, Z.%
, Davis, A.%
, Dean, J.%
\BDBL {}others%
\end{APACrefauthors}%
\unskip\
\newblock
\APACrefYearMonthDay{2016}{}{}.
\newblock
{\BBOQ}\APACrefatitle {Tensorflow: A system for large-scale machine learning}
  {Tensorflow: A system for large-scale machine learning}.{\BBCQ}
\newblock
\BIn{} \APACrefbtitle {12th $\{$USENIX$\}$ symposium on operating systems
  design and implementation ($\{$OSDI$\}$ 16)} {12th $\{$USENIX$\}$ symposium
  on operating systems design and implementation ($\{$OSDI$\}$ 16)}\ (\BPGS\
  265--283).
\PrintBackRefs{\CurrentBib}

\bibitem [\protect \citeauthoryear {%
Arias~Chao%
, Kulkarni%
, Goebel%
\BCBL {}\ \BBA {} Fink%
}{%
Arias~Chao%
\ \protect \BOthers {.}}{%
{\protect \APACyear {2021}}%
}]{%
arias2021aircraft}
\APACinsertmetastar {%
arias2021aircraft}%
\begin{APACrefauthors}%
Arias~Chao, M.%
, Kulkarni, C.%
, Goebel, K.%
\BCBL {}\ \BBA {} Fink, O.%
\end{APACrefauthors}%
\unskip\
\newblock
\APACrefYearMonthDay{2021}{}{}.
\newblock
{\BBOQ}\APACrefatitle {Aircraft Engine Run-to-Failure Dataset under Real Flight
  Conditions for Prognostics and Diagnostics} {Aircraft engine run-to-failure
  dataset under real flight conditions for prognostics and diagnostics}.{\BBCQ}
\newblock
\APACjournalVolNumPages{Data}{6}{1}{5}.
\PrintBackRefs{\CurrentBib}

\bibitem [\protect \citeauthoryear {%
Babu%
, Zhao%
\BCBL {}\ \BBA {} Li%
}{%
Babu%
\ \protect \BOthers {.}}{%
{\protect \APACyear {2016}}%
}]{%
babu2016deep}
\APACinsertmetastar {%
babu2016deep}%
\begin{APACrefauthors}%
Babu, G\BPBI S.%
, Zhao, P.%
\BCBL {}\ \BBA {} Li, X\BHBI L.%
\end{APACrefauthors}%
\unskip\
\newblock
\APACrefYearMonthDay{2016}{}{}.
\newblock
{\BBOQ}\APACrefatitle {Deep convolutional neural network based regression
  approach for estimation of remaining useful life} {Deep convolutional neural
  network based regression approach for estimation of remaining useful
  life}.{\BBCQ}
\newblock
\BIn{} \APACrefbtitle {International conference on database systems for
  advanced applications} {International conference on database systems for
  advanced applications}\ (\BPGS\ 214--228).
\PrintBackRefs{\CurrentBib}

\bibitem [\protect \citeauthoryear {%
Gulli%
\ \BBA {} Pal%
}{%
Gulli%
\ \BBA {} Pal%
}{%
{\protect \APACyear {2017}}%
}]{%
gulli2017deep}
\APACinsertmetastar {%
gulli2017deep}%
\begin{APACrefauthors}%
Gulli, A.%
\BCBT {}\ \BBA {} Pal, S.%
\end{APACrefauthors}%
\unskip\
\newblock
\APACrefYear{2017}.
\newblock
\APACrefbtitle {Deep learning with Keras} {Deep learning with keras}.
\newblock
\APACaddressPublisher{}{Packt Publishing Ltd}.
\PrintBackRefs{\CurrentBib}

\bibitem [\protect \citeauthoryear {%
LeCun%
, Haffner%
, Bottou%
\BCBL {}\ \BBA {} Bengio%
}{%
LeCun%
\ \protect \BOthers {.}}{%
{\protect \APACyear {1999}}%
}]{%
lecun1999object}
\APACinsertmetastar {%
lecun1999object}%
\begin{APACrefauthors}%
LeCun, Y.%
, Haffner, P.%
, Bottou, L.%
\BCBL {}\ \BBA {} Bengio, Y.%
\end{APACrefauthors}%
\unskip\
\newblock
\APACrefYearMonthDay{1999}{}{}.
\newblock
{\BBOQ}\APACrefatitle {Object recognition with gradient-based learning} {Object
  recognition with gradient-based learning}.{\BBCQ}
\newblock
\BIn{} \APACrefbtitle {Shape, contour and grouping in computer vision} {Shape,
  contour and grouping in computer vision}\ (\BPGS\ 319--345).
\newblock
\APACaddressPublisher{}{Springer}.
\PrintBackRefs{\CurrentBib}

\bibitem [\protect \citeauthoryear {%
H.~Li%
, Zhao%
, Zhang%
\BCBL {}\ \BBA {} Zio%
}{%
H.~Li%
\ \protect \BOthers {.}}{%
{\protect \APACyear {2020}}%
}]{%
li2020remaining}
\APACinsertmetastar {%
li2020remaining}%
\begin{APACrefauthors}%
Li, H.%
, Zhao, W.%
, Zhang, Y.%
\BCBL {}\ \BBA {} Zio, E.%
\end{APACrefauthors}%
\unskip\
\newblock
\APACrefYearMonthDay{2020}{}{}.
\newblock
{\BBOQ}\APACrefatitle {Remaining useful life prediction using multi-scale deep
  convolutional neural network} {Remaining useful life prediction using
  multi-scale deep convolutional neural network}.{\BBCQ}
\newblock
\APACjournalVolNumPages{Applied Soft Computing}{89}{}{106113}.
\PrintBackRefs{\CurrentBib}

\bibitem [\protect \citeauthoryear {%
X.~Li%
, Ding%
\BCBL {}\ \BBA {} Sun%
}{%
X.~Li%
\ \protect \BOthers {.}}{%
{\protect \APACyear {2018}}%
}]{%
li2018remaining}
\APACinsertmetastar {%
li2018remaining}%
\begin{APACrefauthors}%
Li, X.%
, Ding, Q.%
\BCBL {}\ \BBA {} Sun, J\BHBI Q.%
\end{APACrefauthors}%
\unskip\
\newblock
\APACrefYearMonthDay{2018}{}{}.
\newblock
{\BBOQ}\APACrefatitle {Remaining useful life estimation in prognostics using
  deep convolution neural networks} {Remaining useful life estimation in
  prognostics using deep convolution neural networks}.{\BBCQ}
\newblock
\APACjournalVolNumPages{Reliability Engineering \& System
  Safety}{172}{}{1--11}.
\PrintBackRefs{\CurrentBib}

\bibitem [\protect \citeauthoryear {%
Liaw%
\ \protect \BOthers {.}}{%
Liaw%
\ \protect \BOthers {.}}{%
{\protect \APACyear {2018}}%
}]{%
liaw2018tune}
\APACinsertmetastar {%
liaw2018tune}%
\begin{APACrefauthors}%
Liaw, R.%
, Liang, E.%
, Nishihara, R.%
, Moritz, P.%
, Gonzalez, J\BPBI E.%
\BCBL {}\ \BBA {} Stoica, I.%
\end{APACrefauthors}%
\unskip\
\newblock
\APACrefYearMonthDay{2018}{}{}.
\newblock
{\BBOQ}\APACrefatitle {Tune: A research platform for distributed model
  selection and training} {Tune: A research platform for distributed model
  selection and training}.{\BBCQ}
\newblock
\APACjournalVolNumPages{arXiv preprint arXiv:1807.05118}{}{}{}.
\PrintBackRefs{\CurrentBib}

\bibitem [\protect \citeauthoryear {%
Peng%
, Wang%
\BCBL {}\ \BBA {} Zi%
}{%
Peng%
\ \protect \BOthers {.}}{%
{\protect \APACyear {2018}}%
}]{%
peng2018switching}
\APACinsertmetastar {%
peng2018switching}%
\begin{APACrefauthors}%
Peng, Y.%
, Wang, Y.%
\BCBL {}\ \BBA {} Zi, Y.%
\end{APACrefauthors}%
\unskip\
\newblock
\APACrefYearMonthDay{2018}{}{}.
\newblock
{\BBOQ}\APACrefatitle {Switching state-space degradation model with recursive
  filter/smoother for prognostics of remaining useful life} {Switching
  state-space degradation model with recursive filter/smoother for prognostics
  of remaining useful life}.{\BBCQ}
\newblock
\APACjournalVolNumPages{IEEE Transactions on Industrial
  Informatics}{15}{2}{822--832}.
\PrintBackRefs{\CurrentBib}

\bibitem [\protect \citeauthoryear {%
Saxena%
, Goebel%
, Simon%
\BCBL {}\ \BBA {} Eklund%
}{%
Saxena%
\ \protect \BOthers {.}}{%
{\protect \APACyear {2008}}%
}]{%
saxena2008damage}
\APACinsertmetastar {%
saxena2008damage}%
\begin{APACrefauthors}%
Saxena, A.%
, Goebel, K.%
, Simon, D.%
\BCBL {}\ \BBA {} Eklund, N.%
\end{APACrefauthors}%
\unskip\
\newblock
\APACrefYearMonthDay{2008}{}{}.
\newblock
{\BBOQ}\APACrefatitle {Damage propagation modeling for aircraft engine
  run-to-failure simulation} {Damage propagation modeling for aircraft engine
  run-to-failure simulation}.{\BBCQ}
\newblock
\BIn{} \APACrefbtitle {2008 international conference on prognostics and health
  management} {2008 international conference on prognostics and health
  management}\ (\BPGS\ 1--9).
\PrintBackRefs{\CurrentBib}

\bibitem [\protect \citeauthoryear {%
B.~Zhao%
, Lu%
, Chen%
, Liu%
\BCBL {}\ \BBA {} Wu%
}{%
B.~Zhao%
\ \protect \BOthers {.}}{%
{\protect \APACyear {2017}}%
}]{%
zhao2017convolutional}
\APACinsertmetastar {%
zhao2017convolutional}%
\begin{APACrefauthors}%
Zhao, B.%
, Lu, H.%
, Chen, S.%
, Liu, J.%
\BCBL {}\ \BBA {} Wu, D.%
\end{APACrefauthors}%
\unskip\
\newblock
\APACrefYearMonthDay{2017}{}{}.
\newblock
{\BBOQ}\APACrefatitle {Convolutional neural networks for time series
  classification} {Convolutional neural networks for time series
  classification}.{\BBCQ}
\newblock
\APACjournalVolNumPages{Journal of Systems Engineering and
  Electronics}{28}{1}{162--169}.
\PrintBackRefs{\CurrentBib}

\bibitem [\protect \citeauthoryear {%
Z.~Zhao%
, Liang%
, Wang%
\BCBL {}\ \BBA {} Lu%
}{%
Z.~Zhao%
\ \protect \BOthers {.}}{%
{\protect \APACyear {2017}}%
}]{%
zhao2017remaining}
\APACinsertmetastar {%
zhao2017remaining}%
\begin{APACrefauthors}%
Zhao, Z.%
, Liang, B.%
, Wang, X.%
\BCBL {}\ \BBA {} Lu, W.%
\end{APACrefauthors}%
\unskip\
\newblock
\APACrefYearMonthDay{2017}{}{}.
\newblock
{\BBOQ}\APACrefatitle {Remaining useful life prediction of aircraft engine
  based on degradation pattern learning} {Remaining useful life prediction of
  aircraft engine based on degradation pattern learning}.{\BBCQ}
\newblock
\APACjournalVolNumPages{Reliability Engineering \& System
  Safety}{164}{}{74--83}.
\PrintBackRefs{\CurrentBib}

\end{thebibliography}

\section*{Appendix}

\subsection*{Hardware and software}

The framework used in this work has been developed with Python 3.7. For network design and training, Tensorflow 2.0 \cite{abadi2016tensorflow} and Keras 2.3 \cite{gulli2017deep} have been used. As model optimization framework, Tune 1.6.0 \cite{liaw2018tune} was selected.

The hardware used was a server with two GTX 1080Ti GPUs, 32GB of RAM memory, and 4 threads. The two GPUs were used to distribute different network training.




\end{document}